\documentclass[10pt,twocolumn,letterpaper]{article}

\usepackage{times}
\usepackage{epsfig}
\usepackage{graphicx}
\usepackage{amsmath}
\usepackage{amssymb}
\usepackage{mathrsfs}
\usepackage{booktabs}
\usepackage{float}
\usepackage{multicol}
\usepackage{multirow}
\usepackage{makecell}

\def\abstract
   {%
   \centerline{\large\bf Abstract}%
   \vspace*{12pt}%
   \it%
   }

\usepackage[pagebackref=true,breaklinks=true,letterpaper=true,colorlinks,bookmarks=false]{hyperref}

\newcommand{\be}{\begin{equation}}
\newcommand{\ee}{\end{equation}}
\newcommand{\ba}{\begin{array}}
\newcommand{\ea}{\end{array}}
\newcommand{\bea}{\begin{eqnarray}}
\newcommand{\eea}{\end{eqnarray}}
\newcommand{\balg}{\begin{align}}
\newcommand{\ealg}{\end{align}}
\newcommand{\bit}{\begin{itemize}}
\newcommand{\eit}{\end{itemize}}
\newcommand{\trm}[1]{\textrm{#1}}

\newcommand{\tbf}[1]{\textbf{#1}}

\newcommand{\rat}[1]{\renewcommand{\arraystretch}{#1}}

\newcommand{\Mpc}{\trm{\Mpc}}
\newcommand{\yr}{\trm{\yr}}
\newcommand{\eV}{\trm{\eV}}

\begin{document}

\title{Rethinking Text Line Recognition Models}

\author{Daniel Hernandez Diaz\\
Google Research\\
{\tt\small dhernandezdiaz@google.com}
\and
Siyang Qin\\
Google Research\\
{\tt\small qinb@google.com}
\and
Reeve Ingle \\
Google Research\\
{\tt\small reeveingle@google.com}
\and
Yasuhisa Fujii \\
Google Research\\
{\tt\small yasuhisaf@google.com}
\and
Alessandro Bissacco \\
Google Research\\
{\tt\small bissacco@google.com}
\\
}

\date{}
\maketitle

\begin{abstract}
In this paper, we study the problem of text line recognition. Unlike most approaches targeting specific domains such as scene-text or handwritten documents, we investigate the general problem of developing a universal architecture that can extract text from any image, regardless of source or input modality. We consider two decoder families (Connectionist Temporal Classification and Transformer) and three encoder modules (Bidirectional LSTMs, Self-Attention, and GRCLs), and conduct extensive experiments to compare their accuracy and performance on
widely used public datasets of scene and handwritten text. We find that a combination that so far has received little attention in the literature, namely a Self-Attention encoder coupled with the CTC decoder, when compounded with an external language model and trained on both public and internal data, outperforms all the others in accuracy and computational complexity. Unlike the more common Transformer-based models, this architecture can handle inputs of arbitrary length, a requirement for universal line recognition. 
Using an internal dataset collected from multiple sources, we also expose the limitations of current public datasets in evaluating the accuracy of line recognizers, as the relatively narrow image width and sequence length distributions do not allow to observe the quality degradation of the Transformer approach when applied to the transcription of long lines.

\end{abstract}

\section{Introduction}

Optical character recognition (OCR) is a crucial component of a wide range of practical applications, such as visual search, document digitization, autonomous vehicles, augmented reality (e.g., visual translation), and can increase the environmental awareness of visually impaired people \cite{neat2019scene}. In the last decade, increasing sophistication in OCR models has come hand-in-hand with a remarkable increase in accuracy for various domains~\cite{4531750,6751207,6628705,long2020scene}.

Traditional approaches usually separate the text extraction task into two subproblems: \emph{text detection} and \emph{text recognition}. Text detection algorithms try to detect text instances (words or lines) in the input images, while text recognition models try to decode the textual content from cropped and rectified text patches. For scene (sparse) texts, the majority of the text spotting methods operate at the \emph{word-level}, driven by the fact that most of the public datasets in the domain only provide word-level labeling. On the other hand, handwriting and dense text recognition systems typically adopt \emph{lines} as the processing unit~\cite{6751207,6628705,reeve}.

\begin{figure}
\begin{center}
  \includegraphics[width=1.0\linewidth]{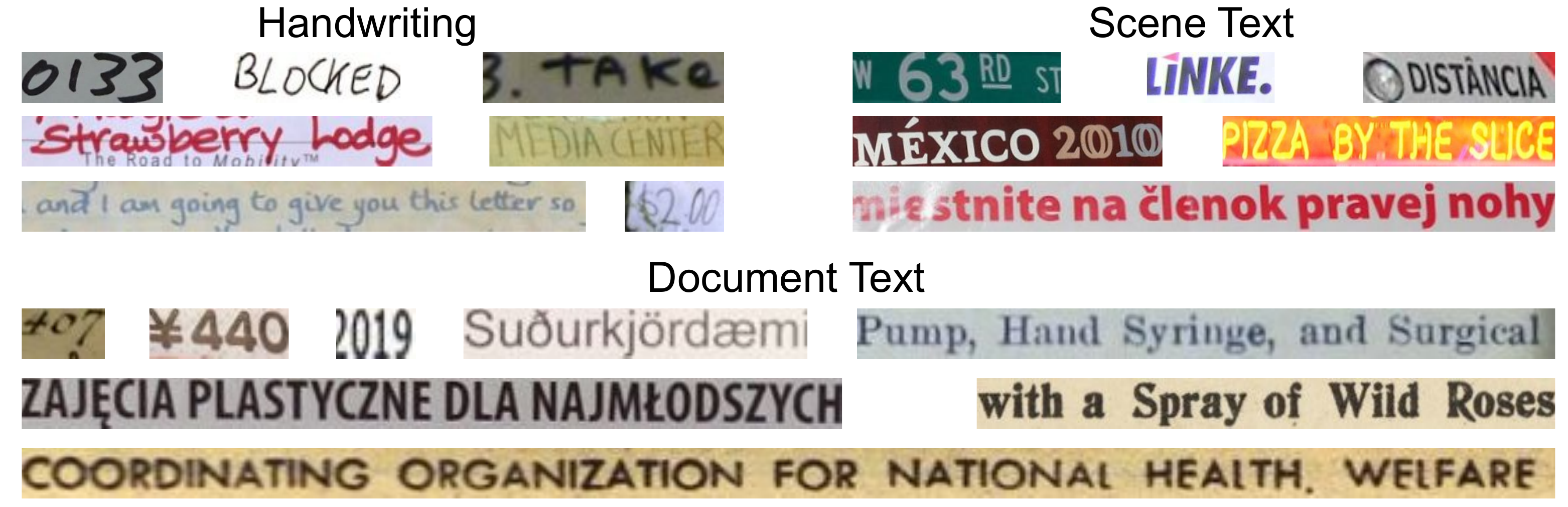}
  \caption{Example text line images in our Internal dataset, which contains handwriting, scene-text, and document text images of various lengths.}
\label{fig:example_images}
\end{center}
\end{figure}


Despite the popularity of word-level systems for scene text recognition, operating at world-level could be suboptimal for a general purpose system supporting many scripts. First, for handwritten text, it is not always straightforward to separate words since detected boxes for neighboring words could partially overlap. Second, for dense printed documents, it is difficult to reliably detect a large amount of small and dense words. Third, it is challenging for detectors to separate words for scripts that do not separate words by spaces such as Chinese, Japanese, and Korean. Last but not least, word-level OCR models are more likely to miss punctuation and diacritic marks.

In this work, we consider a universal \emph{line-based} OCR pipeline that can extract dense printed or handwritten text as well as sparse scene-text with the goal of determining the optimal \emph{text-line recognition} (TLR) model in this scenario. In such a pipeline, a separate the text detection model is responsible for
detecting and rectifying all lines of text in the image. As this step is outside the scope of this paper, we exclude from the analysis here public datasets  with irregular scene text  where detection and rectification are the main focus, such as \cite{totaltext}.
In a line-based OCR system, \emph{the recognizer needs to handle arbitrarily long text-lines effectively and efficiently}. This challenge has been overlooked by previous state-of-the-art word-based models. Public datasets (mostly word level) have relatively narrow image width and sequence length distributions (see Figure \ref{fig:width_distro}). At the same time, the design, performance and speed of TLR models depends on these variables. To understand these issues, we study TLR models trained on an internal dataset containing examples within a wide range of lengths and with a much larger symbol coverage (675 classes containing most of the symbols in Latin script and special characters, compared to alphanumeric characters used in majority of the public datasets). We found that techniques such as \emph{image chunking} substantially alleviate the challenges posed by examples of wildly different sizes. 

We explore a number of architectures for TLR and compare their accuracy and performance on the internal dataset. From this study emerge concrete recommendations for the design of universal, line-level, TLR systems. \emph{We find that the model that uses a Self-Attention encoder \cite{transformer} coupled with the Connectionist Temporal Classification (CTC) decoder \cite{ctc}, compounded with an explicit language model, outperforms all other models having both maximal text-line recognition accuracy and minimal complexity}. To the best of our knowledge, this model architecture has not been previously studied in the literature. It also fits within the overarching trend in machine learning of disposing of recurrent networks in favor of attention modules. More generally, encoder modules based on the self-attention mechanism emerge winners in our encoder comparison, regardless of decoder type. Our performance on widely used public datasets (including printed, handwritten and scene-text) are reported. We also report results of training this model on public datasets only, and show that even when using such a restricted train set accuracy is comparable to the widely used Transformer-based approach, and not far from significantly more complex state-of-the-art models. 

To summarize, our contributions are four-fold:
\begin{itemize}
    \item We fairly compare 9 different model design choices on an internal dataset and on a \emph{universal} TLR task, pairing each of three encoders (Self-Attention, GRCL, and BiLSTM) to each of three decoders (CTC, CTC with LM and the Transformer decoder).
    \item We identify as the optimal model architecture a previously unstudied combination of Self-Attention encoding with CTC decoding. We found that this model yielded the top accuracy with low memory requirements and very good latency, making it an easy choice for low-resource environments.
    \item We propose the use of image chunking to ensure that the model works efficiently and effectively on arbitrary long input images without shrinking.
    \item We find that models based on the Transformer decoder have significant  regression in quality if the image length is outside the length distribution of the training data.
\end{itemize}

\section{Related Work}
\label{related_work}

Handwriting and scene-text are usually treated separately in TLR, and most work in the literature focuses on one or the other. Nevertheless, models for these domains share many features in common. Generally speaking, they are designed around the use of a CNN for extracting a sequence of visual features, that is subsequently passed to an encoder/decoder network that outputs the labels. 

For the case of Scene-Text Recognition (STR), an in-depth survey was recently carried out by Chen et al. \cite{Chen2020TextRI}. Close in spirit to this paper is Baek et al. \cite{Baek2019WhatIW} in the domain of scene-text recognition. That work emphasized the inconsistencies among the public datasets typically used for training and evaluation in the literature, and was the first attempt to provide a uniform framework for the comparison of recognizer models. In their analysis, they include a transformation module, that preprocesses a potentially irregular input image before it is fed to the vision backbone. Compared to that work, this paper is a more focused effort. Assuming that images are rectified and fixing the vision backbone allows us to perform a more straightforward encoder/decoder comparison, also scanning over different encoder configurations. 

CTC models for STR are well studied \cite{7801919, GAO2019161, su_lu}. The seminal work by Shi et al. \cite{7801919} proposed the standard, end-to-end backbone + encoder/decoder architecture for TLR tasks. Our Self-Attention / CTC model replaces the RNN in that work by a Self-Attention module. Sequence-to-Sequence (Seq2Seq) models with attention have become more prevalent for STR in recent years \cite{sahan, GCAM, nrtr_shengetal, bleeker2019bidirectional, lee2020recognizing, GCAM}, in part due to the advent of the Transformer architecture \cite{transformer}, that has allowed the field to loosen its ties to the recurrent connection. Sheng et al. \cite{nrtr_shengetal} first applied the Transformer to STR \cite{bleeker2019bidirectional}. Lee et al. \cite{lee2020recognizing} devised a novel 2D dynamic positional encoding method with impressive results. Other models with an initial rectification step \cite{Shi_2016_CVPR, Baek2019WhatIW, scatter, shietal, esir_zhan_lu} have been most successful in irregular STR. In particular, the drop in accuracy for longer lines was noticed in \cite{shietal}. 
Wang et al \cite{dan_for_tr} combine a fully convolutional (FCN) encoder with an RNN decoder based on Gated Recurrent Units (GRU).

Modern methods for handwriting recognition (HWR) have been reviewed by Memon et al. \cite{9151144}. Michael et al. \cite{8978104} compare different attention mechanisms for Seq2Seq methods in the HWR domain. RNN encoders with CTC decoders have been customary for some time, with Long-Short Term Memory (LSTM) being the typical RNN cell of choice \cite{gcrnn_hwr_bluche_messina, NIPS2008_3449, dropout_improves_rnn_hwr}. Multidimensional LSTM encoders have also been tried, e.g. \cite{NIPS2008_3449, large_multidimensional_lstm_hwr, are_mdrnns_needed_for_htr}. Seq2Seq approaches to handwriting recognition including attention-based decoding have been catching up \cite{8978104, kang2020pay, Poulos2017AttentionNF, 2018arXiv180707965C,  NIPS2016_2bb232c0}. 
The difficulties with scaling handwriting recognition were tackled in \cite{reeve}.

\setlength\belowcaptionskip{-1ex}
\begin{figure*}[t]
\begin{center}
  \includegraphics[width=0.9\linewidth]{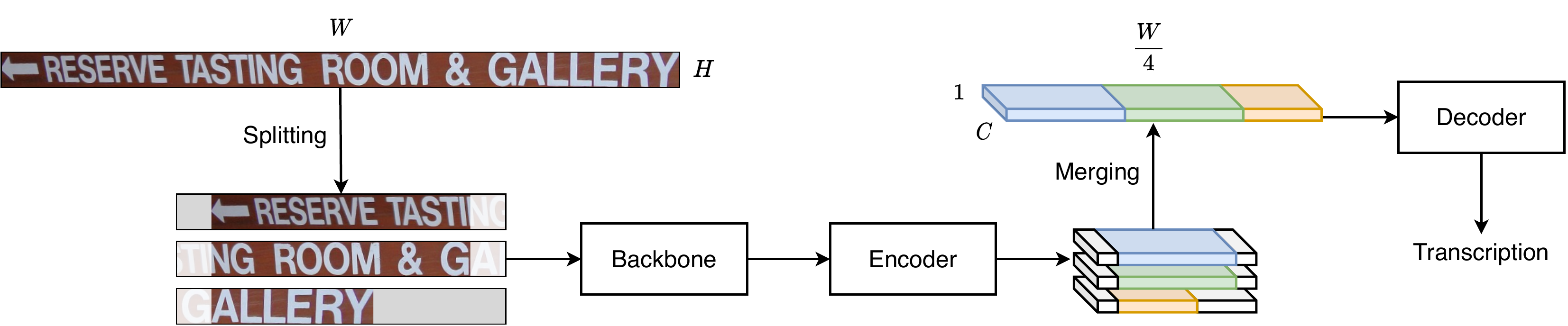}
\end{center}
  \caption{\small This figure shows the image chunking process. Input images are split into overlapping chunks with bidirectional padding before being fed to the backbone. The valid portions of the produced sequence features are concatenated before being fed to the decoder.}
\label{fig:chunking}
\end{figure*}

\section{Anatomy of a Text-Line Recognition Model}
\label{anatomy}

Most state-of-the-art TLR algorithms consists of three major components: a convolutional backbone to extract visual features; a sequential encoder which aggregates features from part or entire sequence; and finally a decoder which produces the final transcription given the encoder output. In this work, we study different combination of encoder and decoder with a fixed backbone and propose an optimal model architecture.

\subsection{Backbone}

Similar to other vision tasks, a convolutional backbone network is used to extract relevant visual features from text-line images. Any of the many mainstream vision modules in the literature can be reused in principle as a TLR backbone; ResNet \cite{resnet_orig}, Inception \cite{inception}, and MobileNet-like \cite{mobilenet, mobilenetv2} networks are standard choices. Generally speaking, increasing the complexity of the backbone translates into moderate performance gains. In this work, we use a fixed backbone across all of the experiments while focusing on the question of finding the optimal encoder/decoder combination.

\begin{figure}[t]
\begin{center}
  \includegraphics[width=1.0\linewidth]{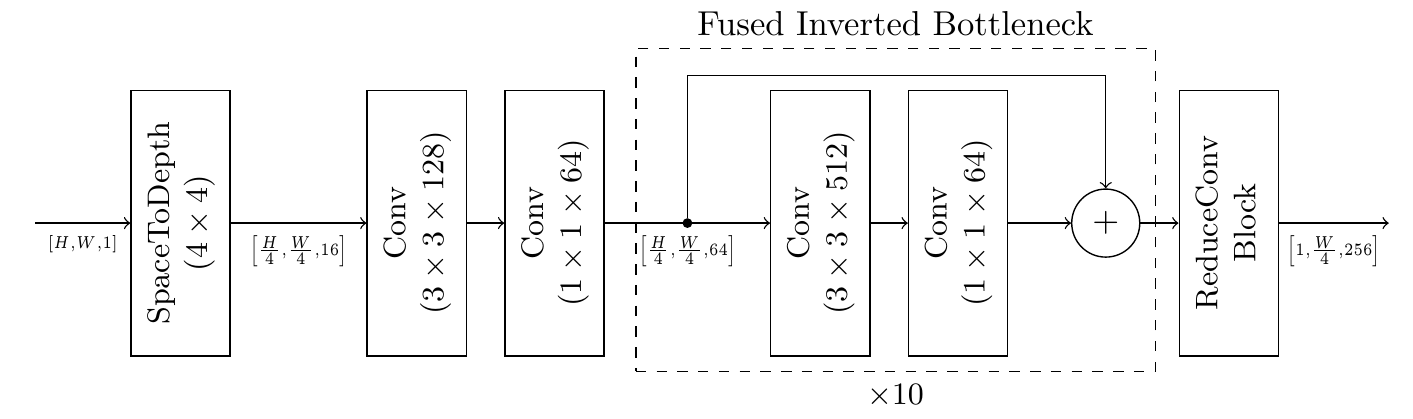}
\end{center}
  \caption{Backbone used in our experiments. We first reduce the resolution of input grayscale image by 4x with a space-to-depth operation, then we apply $11$ Fused Inverted Bottleneck \cite{xiong2020mobiledets} layers with expansion rate $8$ and $64$ output channels, and use a residual convolution block to project the output into a tensor of height 1. 
  }
\label{fig:backbone}
\end{figure}

Our backbone is an isometric architecture \cite{sandler2019nondiscriminative} using Fused Inverted Bottleneck layers as building block, a variant of the Inverted Bottleneck layer \cite{mobilenetv2} that replaces separable with full convolutions for higher model capacity and inference efficiency on modern accelerators \cite{xiong2020mobiledets}. Isometric architectures maintain constant internal resolution throughout all layers, allowing a low activation memory footprint and making it easier to customize the model to specialized hardware and achieve the highest utilization. Figure \ref{fig:backbone} illustrates the network in detail. It consists of a space-to-depth layer with block size 4, followed by 11 Fused Inverted Bottleneck layers with $3\times 3$ kernels and $8\times$ expansion rate with $64$ output channels. A final fully convolutional residual block is applied to reduce the tensor height to 1, to be fed as input to the encoder network~\cite{reeve}.

\subsection{Encoder}

The backbone has a finite receptive field, which limits its ability to encode wide range context information. This makes it difficult to decode long sequences under challenging scenarios (e.g., handwritten text). Standard sequential encoders can be used to capture long-distance context features, making it a crucial component of the TLR model. In this work, we compare the following encoders:

\tbf{BiLSTM}: Bidirectional RNNs based on the  LSTM cell \cite{lstm} are natural candidates to capture long-range correlations in the output feature sequence of the backbone. In our comparison, we stack 1 to 3 LSTM layers, with each cell having 512 hidden units.

\begin{figure}[t]
\begin{center}
  \includegraphics[width=1.0\linewidth]{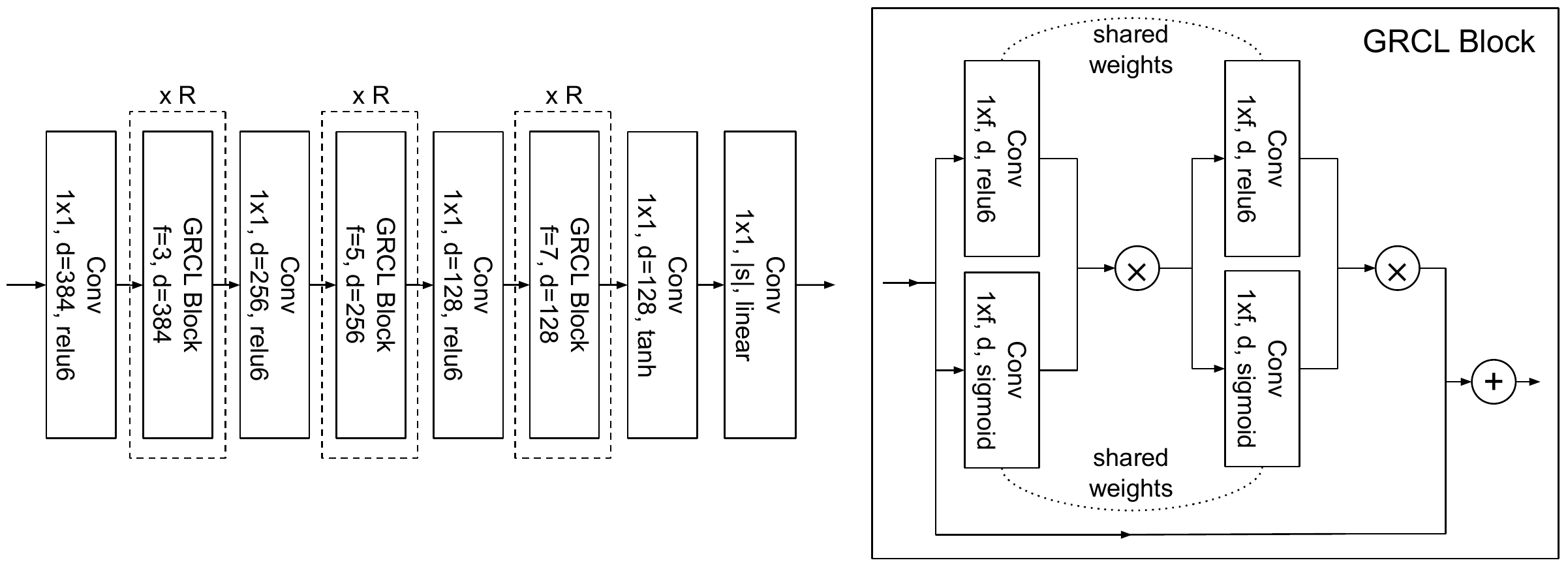}
\end{center}
  \caption{The GRCL encoder architecture.}
\label{fig:grcl_encoder}
\end{figure}

\tbf{Gated Recurrent Convolution Layer (GRCL)}: GRCL was introduced in \cite{NIPS2017_6637} and has demonstrated its effectiveness on OCR. Our GRCL encoder consists of 3 sets of repeated GRCL blocks. The convolutions in each of the 3 sets use [384, 256, 128] output filters and kernel widths (1-D convolution along width dimension) [3, 5, 7] respectively (Fig.~\ref{fig:grcl_encoder}). The basic layer in the repeated blocks of each set contains a convolutional gating mechanism, which uses a convolution with sigmoid activation to control information propagation from a convolution with ReLU activation. We compare 6 different variations where the number of GRCL blocks within each set varied from 1 to 6. Batch Renormalization \cite{batch_renormalization} was used for each of the blocks and a dropout rate of 0.1 was applied in all the convolutional layers. 

\tbf{Self-Attention}: The Self-Attention encoder, proposed in Vaswani et al. \cite{transformer} has been widely used in numerous NLP and vision tasks. As an image-to-sequence task, text-line recognition is no exception. The Self-Attention encoder can effectively output features that summarize an entire sequence without making use of recurrent connections. The output of the backbone, with height dimension being removed ($X\in \mathbb{R}^{n\times d}$), is fed to the encoder. The encoded feature $Y$ is computed as:

\begin{equation}
\begin{gathered}
Q=XW_q\,,\quad K=XW_k\,,\quad V=XW_v \,,\\
Y=\textrm{softmax}\left( \frac{QK^T}{\sqrt{d}} \right)V \,,
\end{gathered}
\end{equation}
where $W_q$, $W_k$ and $W_v$ are $d\times d$ learned parameters which project the input sequence $X$ into queries, keys and values respectively. The encoded feature $Y$ is a convex combination of the computed values $V$, the similarity matrix is computed by the dot-product of queries and keys.

\begin{figure*}[t]
\begin{center}
   \includegraphics[width=0.95\linewidth]{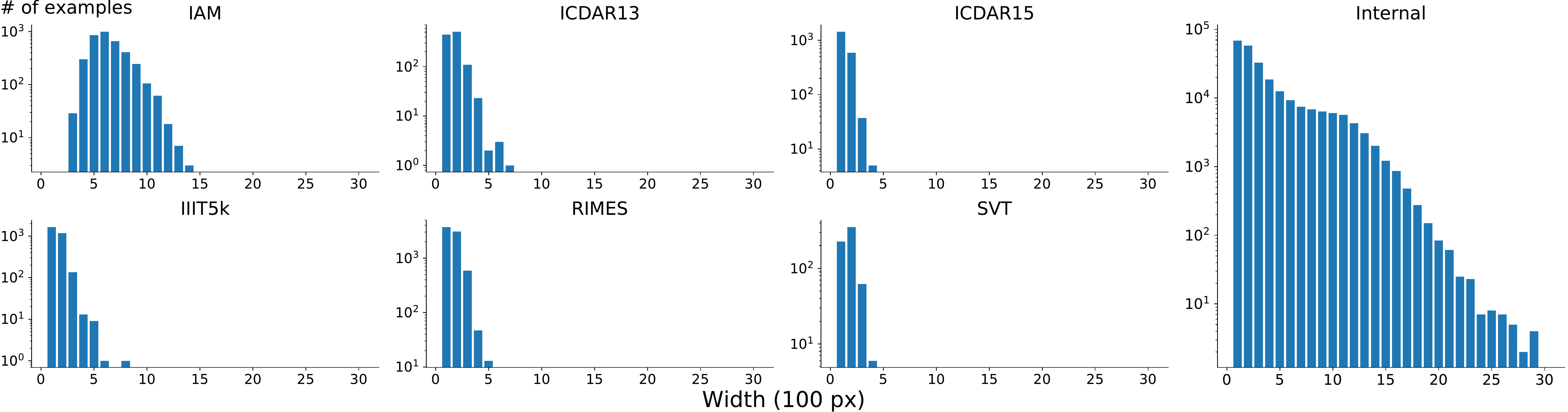}
\end{center}
   \caption{Height-normalized (40 px) distribution of text line images widths for the different datasets used in this work.}
\label{fig:width_distro}
\end{figure*}

We use multi-head attention with 4 separate heads. The hidden size is set to 256. To prevent over-fitting, we apply dropout after each sub-layer with drop ratio 0.1. Sinusoidal relative positional encoding is added to make the encoder position-aware. In our experiments, we compare the accuracy and complexity of different model variations by stacking $k$ encoder layers with $k\in \{4, 8, 12, 16, 20\}$.

\subsection{Decoder}

Decoders take the encoded sequence features and try to decode its text content. CTC \cite{ctc} and attention decoders are among the popular choices for text-line recognition. In this work, we compare the CTC decoder (with and without language model) and the Transformer decoder \cite{transformer}.

\tbf{CTC}: The CTC decoder was originally used in speech recognition, and researchers have had huge success adopting it to OCR tasks. The input encoded features can be viewed as vertical feature frames along the width dimension. The input is fed to a dense layer to obtain per-frame probabilities over possible labels. At inference time, we use a simple greedy CTC decoding strategy which removes all repeated symbols and all blanks from the per-frame predictions. Beam search decoding can be optionally used to consider multiple valid paths which are mapped to the same final output sequence. Our empirical study shows that beam search is not needed when CTC is paired with a powerful encoder such as the self-attention encoder.

\tbf{CTC w/ LM}: Incorporating an explicit language model on top of the logits from the optical model can significantly improve accuracy. We follow the approach described in~\cite{fujii:icdar2017} to form a log-linear model, combining the cost of CTC logits, the cost of a character-based $N$-gram language model, the cost of a character unigram prior, and transition costs defined for new characters, blank labels, and repeated characters. The weights of the feature functions were optimized using minimum error rate training~\cite{macherey:emnlp2008}. 

\tbf{Transformer Decoder}: The Transformer decoder has become the decoder of choice for sequence prediction tasks such as machine translation. In our experiments, we incorporate a standard Transformer decoder with a stack of 8 repeated layers. In addition to self-attention and feed-forward layers, the decoder adds a third sub-layer which applies single-head attention over the encoded features (cross-attention to compute a context vector). The decoder's hidden size is set to 256. Greedy decoding is applied at inference time. Per-step cross-entropy loss with label smoothing is used during training.

\subsection{Chunking}
\label{sec:chunking}

Due to the dot-product attention in the self-attention layer, the model complexity and memory footprint of the encoder grows quadratically as a function of the image width. This can cause issues for inputs of large width. Shrinking wide images can avoid such issues but it will inevitably affect recognition accuracy, especially for narrow or closely-spaced characters.

We propose a simple yet effective chunking strategy to ensure that the model works well on arbitrarily wide input images without shrinking (see Figure \ref{fig:chunking}), similar to~\cite{KaiChen:ICDAR2015}. We resize the input image to 40 pixels height, preserving the aspect ratio. Then the text-line is split into overlapping chunks with bidirectional padding to reduce possible boundary effects (note the last chunk has extra padding to ensure a uniform shape for batching purpose). We feed overlapping chunks into the backbone and self-attention encoder to produce sequence features for each chunk. Finally, we merge the valid regions back into a full sequence, removing the padding areas.

This approach splits long sequences into $k$ shorter chunks, effectively reducing model complexity and memory usage by a factor of $k$ for self-attention layers. This strategy is used both at training and inference time to keep a consistent behavior. Via a controlled experiment, we find no regression by applying chunking to the inputs. The aspect ratio of each chunk is set to 8 (i.e., 320 pixel width for 40 pixel height) to provide a reasonable amount of context information.


\section{Experimental Setup}
\label{experimental_setup}

\begin{table*}[t]
\begin{center}
{\footnotesize
\centering
    \caption{Evaluation results on public handwriting and scene-text datasets of our best models and selected works. The ``Rect.'' column indicates whether the model includes a rectification module. ``S-Attn'', ``Attn'', and ``Tfmr Dec.'' stand for Self-Attention, Attention and Transformer Decoder respectively. ``MJ'', ``ST'' and ``SA'' stand for MJSynth\cite{synth90ka, synth90kb}, SynthText \cite{synthtext} and SynthAdd \cite{Li_Wang_Shen_Zhang_2019} respectively.}
\rat{1.2}
    \begin{tabular}{@{}c|ccc|cccccc@{}}\toprule
    & Rect. & Encoder / Decoder & Train Dataset & IAM $\downarrow$ & RIMES $\downarrow$ & IIIT5K $\uparrow$ & SVT $\uparrow$ & IC13 $\uparrow$ & IC15 $\uparrow$ \\
    \hline
    Bluche and Messina \cite{gcrnn_hwr_bluche_messina} & No & GCRNN/CTC & IAM (50k lexicon) & 3.2 & \tbf{1.9} & - & - & - & - \\
    Michael et al. \cite{8978104} & No & LSTM/LSTM w/Attn & IAM & 4.87 & - & - & - & - & - \\
    Kang et al. \cite{kang2020pay} & No & Transformer & IAM & 4.67 & - & - & - & - & - \\
    Bleeker and de Rijke \cite{bleeker2019bidirectional} & No & Transformer & MJ + ST & - & - & 94.7 & 89.0 & 93.4 & 75.7 \\ 
    Lu et al. \cite{master} & No & \thead{Global Context Attn \\ / Tfmr Dec.} & MJ + ST + SA & - & - & 95.0 & 90.6 & 95.3 & 79.4 \\     
    Qiao et al. \cite{qiao2020gaussian} & No & LSTM / LSTM + Attn & MJ + ST & - & - & 94.4 & 90.1 & 93.3 & 77.1 \\
    Sheng et al. \cite{nrtr_shengetal} & Yes & S-Attn / Attn & MJ + ST & - & - & 93.4 & 89.5 & 91.8 & 76.1 \\
    Shi et al. \cite{shietal} & Yes & LSTM / LSTM + Attn & MJ + ST & - & - & 91.93 & 93.49 & 89.75 & - \\
    Wang et al. \cite{dan_for_tr} & No & FCN / GRU & MJ + ST & 6.4 & 2.7 & 94.3 & 89.2 & 93.9 & 74.5 \\
    SCATTER \cite{scatter} & Yes & CNN / BiLSTM & MJ + ST + SA & - & - & 93.9 & 92.7 & 94.7 & \tbf{82.8} \\
    Yu et al. \cite{yu2020towards} & No & \thead{Attn / \\ Semantic Attn.}  & MJ + ST  & - & - & 94.8 & 91.5 & 95.5 & 82.7 \\
    \hline
    \multirow{9}{*}{Ours} & No & S-Attn / CTC & MJ + ST + Public & 
    -  & - & 92.06 & 87.94 & 92.15 & 72.36 \\
    & No & S-Attn / CTC + LM & MJ + ST + Public  & -  & - & 93.63 & 91.50 & 93.61 & 74.77 \\
    & No & Transformer & MJ + ST + Public & - & - & 93.93 & 92.27 & 93.88 & 77.08 \\
    
    & No & S-Attn / CTC & Internal & 
    4.62  & 10.80 & 96.26 & 91.96 & 94.43 & 78.43 \\
    & No & S-Attn / CTC + LM & Internal  & 3.15  & 7.79 & \tbf{96.83} & \tbf{94.59} & \tbf{95.98} & 80.36 \\
    & No & Transformer & Internal & 3.99 & 9.71 & 96.54 & 92.59 & 94.34 & 79.68 \\ 
    
    & No & S-Attn / CTC & Internal + Public & 3.53  & 2.48 & 95.66 & 91.34 & 93.70 & 78.38 \\
    & No & S-Attn / CTC + LM & Internal + Public & \tbf{2.75}  & 1.99 & 96.43 & 93.66 & 95.25 & 79.92 \\    
    & No & Transformer & Internal + Public & 2.96 & 2.01 & 96.44 & 92.50 & 93.97 & 80.45 \\
    \bottomrule
    \end{tabular}
    \label{tab:sota-comparison}
}
\end{center}
\end{table*}

\subsection{Datasets}
\label{datasets}

To better train and evaluate TLR models for a line-based OCR engine with broad language support, we built an \tbf{Internal dataset} which contains both single words and long text-lines (see Figure \ref{fig:width_distro}). 
This dataset has multiple image sources, including synthetic text, scene-text, handwritten, documents, and photos reflecting some common phone camera use cases (translate, receipts, notes and memos). Scene-text training consists of the Uber-Text \cite{UberText} and the OpenImages dataset \cite{Kuznetsova_2020} self-annotated with Google Cloud Vision TEXT\_DETECTION API \cite{CloudVisionOcr}. Other images are from public sources - Flickr, Web pages - or collected through vendors specialized in crowdsourcing, which were given detailed instructions on how to collect the images to mimic photos taken by phone application users. The images contain text from many Latin languages, with 675 symbols including punctuation and special characters, and were annotated by human raters with a transcription/verification iterative process seeded from the OCR predictions. 

We also evaluate our models on six commonly used public benchmarks in the scene-text and handwriting domains, namely, IAM and RIMES for handwritten text, and IIIT5K, SVT, ICDAR13 and ICDAR 15 for scene-text  (detailed descriptions of these datasets can be found in the Supplementary Material). Despite their wide usage, public datasets share two features that make them somewhat unrealistic, namely, \emph{relatively narrow image-length distribution} and \emph{lack of symbol coverage}. The height-normalized image width distribution of the test set of the public datasets is shown in the left panels in Figure \ref{fig:width_distro}. The IAM dataset lacks short lines while in the other word-level datasets long lines are missing.



\subsection{Implementation Details}


Models were trained on three different training data configurations : \emph{Synth + Public}, \emph{Internal}, and \emph{Internal + Public}. Most of the analysis presented below corresponds to training on the \emph{Internal} configuration, where only the internal dataset was used for training. To establish a qualitative comparison with other works, we also trained on the \emph{Internal + Public} configuration, where data from \emph{Internal} was combined with the training data from the public datasets (\emph{Public}). These two were then mixed in a 9:4 ratio.

 The \emph{Synth + Public} runs are scene-text, public-data-only runs, that combine a merge of MJSynth\cite{synth90ka, synth90kb} and Synthtext\cite{synthtext} with the training data from the public datasets. The latter was further augmented 10 times by applying random distortions (affine, blur, noise, etc). During training, \emph{synth} and \emph{train} were mixed in a 10:3 ratio.

All of our models were trained and tested using grayscale images. Image chunking (Section \ref{sec:chunking}) is applied to \emph{CTC based models} at both training and inference time to efficiently handle arbitrarily long inputs. We normalize the image height to 40 pixels while keeping the aspect ratio. For models which use the \emph{Transformer decoder}, to prevent accuracy regression on long lines (Section \ref{resizing}), we feed fixed size images (resized and padded to $1024\times 40$) to the models. Since the image width is bounded, chunking strategy is not used for \emph{Transformer decoder} based models.

The character-based $N$-gram language model for the CTC w/ LM decoder was trained from Web texts with stupid-backoff~\cite{TBrants:EMNLP2007} and $N$ was set to 9. The weights of the feature functions were tuned on the development portion of the internal dataset. 



All the models are implemented in TensorFlow \cite{abadi2016tensorflow}, trained using 32 TPU-v3 cores with a batch size of 1024. The SGD optimizer is used with momentum set to 0.9. We use a linear learning rate warm-up followed by stepped learning rate decay. Training iterations, initial learning rate, weight decay factor and global gradient clipping ratio are optimized for each model using grid search.

\subsection{Evaluation Protocol}

The handwriting datasets IAM and RIMES are evaluated using the case-sensitive Character Error Rate (CER), defined as the Levenshtein distance at the character level between the prediction and the ground truth, normalized by the ground truth length. For the scene-text public datasets, we use case-insensitive word prediction accuracy (WPA) to evaluate the  recognition results as is customary. This is defined as $1 - W$ where $W$ is the Word Error Rate, defined through the Levenshtein distance, analogously to CER, but at the word level.

Evaluation on the internal test set, that includes images from the printed, handwritten and scene-text domains, is evaluated using both WPA and CER. No lexicon is used in any of our experiments.

\section{Result Analysis}
\label{results}

\subsection{Encoder / Decoder Comparisons}

\begin{figure*}[t]
\begin{center}
\includegraphics[width=0.95\linewidth]{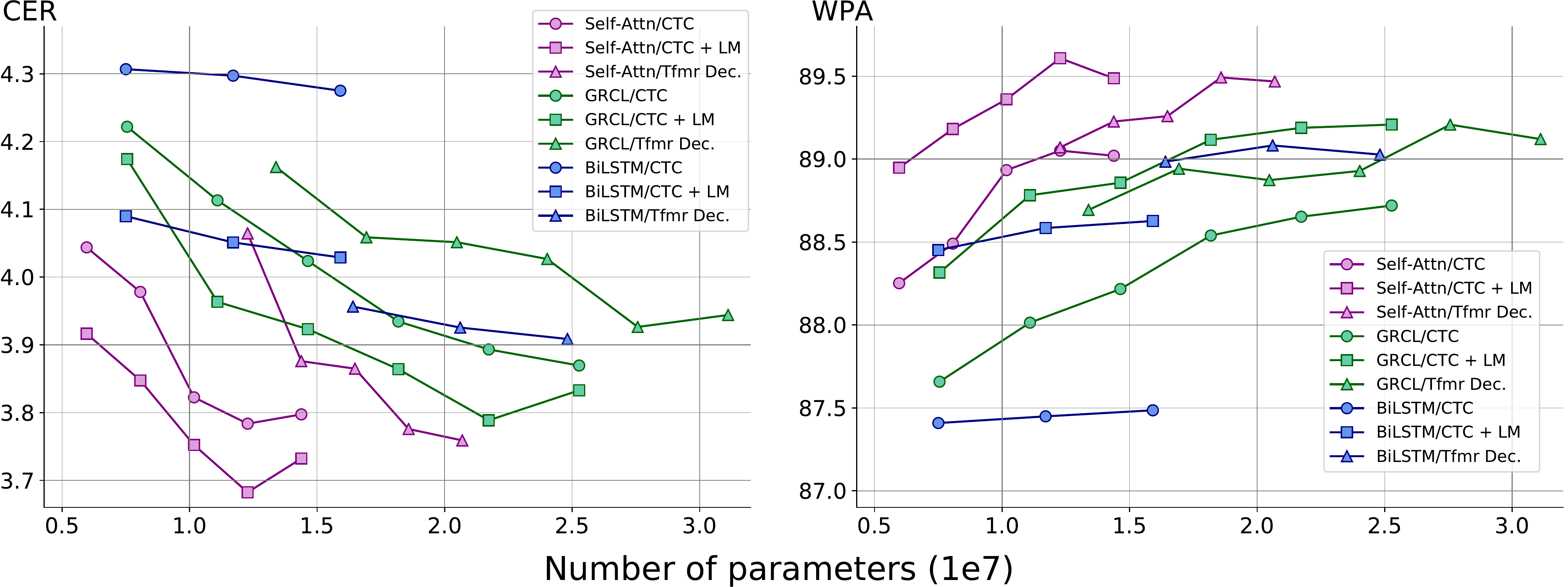}
\end{center}
  \caption{CERs and WPAs for the different encoder/decoder combinations evaluated on the internal test set. \emph{Self-Attention:} 5 models are plotted per decoder corresponding to 4, 8, 12, 16, and 20 Self-Attention layers. \emph{GRCL:} 6 models are plotted corresponding to the number of GRCL blocks per set, from 1 to 6. \emph{BiLSTM:} Three models are plotted with increasing depth 1, 2, and 3. }
\label{fig:comparison_cers}
\end{figure*}

Evaluation results on the \emph{internal} test set are shown in Fig.~\ref{fig:comparison_cers} for CER and WPA, respectively. Within each figure, line color is used to indicate the encoder type: Self-Attention (pink), GRCL (green), or BiLSTM (blue). Alternatively, decoder type is indicated by the symbol: CTC (circle), CTC with LM (square), or Transformer Decoder (triangle).

\setlength\belowcaptionskip{-1ex}
\setlength\belowcaptionskip{-3ex}
\begin{figure}[h]
  \includegraphics[width=0.95\linewidth]{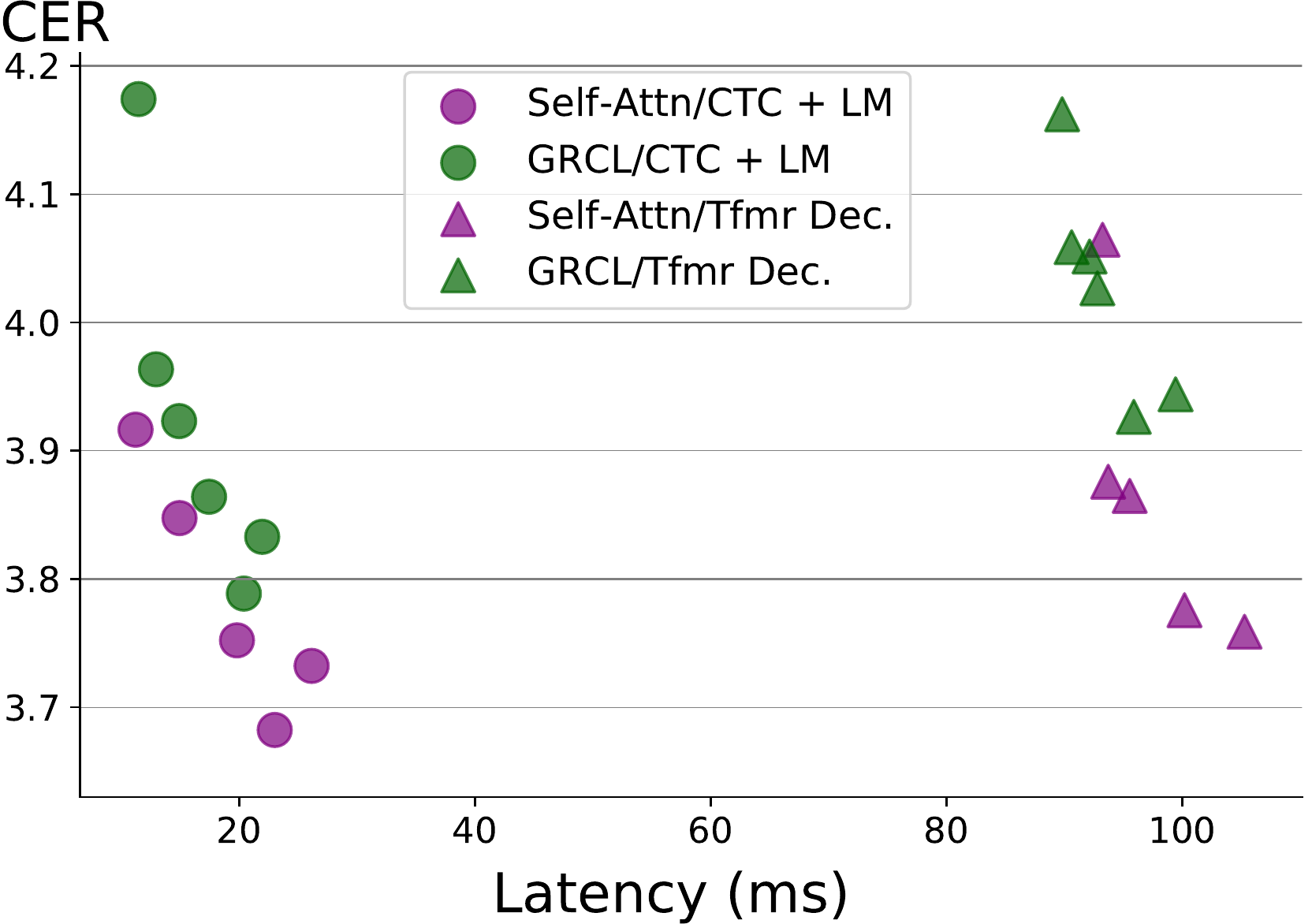}
\caption{CER vs Latency for the non-recurrent models in Fig.~\ref{fig:comparison_cers}. CTC decoding is 5-10 times faster. The LM adds a small overhead (1 ms). For all models, latency was computed on a NVIDIA V100 GPU card. } 
\label{fig:latencies}
\end{figure}

The Self-Attention encoder (pink lines) is the most performant encoder irrespective of the decoder. With similar number of parameters, the CER for models with the Self-Attention encoder was $\sim$ 2.5 - 10\% lower (relative) than for their GRCL and BiLSTM counterparts, translating into WPA gains of $\sim$ 1\% absolute. Among all the models considered, the models with best CER and WPA for all three decoders were Self-Attention models. 

CTC decoders compounded with an explicit LM achieve the best CER and WPA for the Self-Attention and GRCL encoders. The CER gains afforded by the LM applied to each of the CTC models considered, were relatively modest ($<$5\% relative for fixed number of parameters for Self-Attention and GRCL). There is a bigger improvement from the LM for the BiLSTM encoder, but in that case, the Transformer Decoder outperformed CTC, an indication of the importance of the Attention mechanism in RNN architectures. 

\begin{figure*}[t]
\begin{center}
  \includegraphics[width=0.95\linewidth]{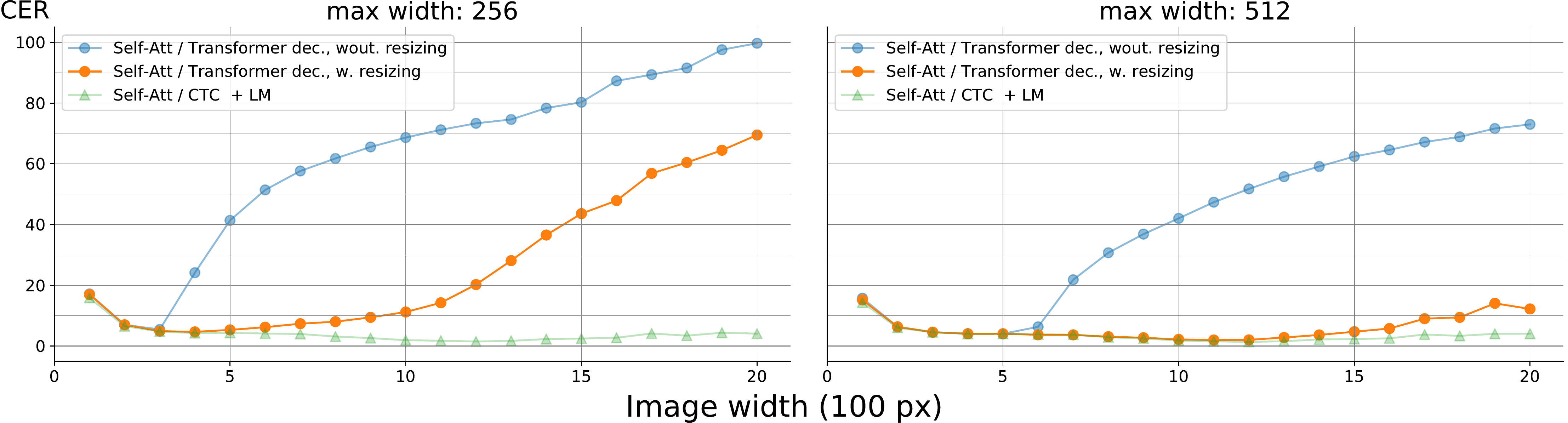}
\end{center}
  \caption{CER per 100 px bucket of height-normalized image width showing the impact of resizing during inference in the Transformer architecture trained with images up to  256px wide (left) and up to 512px wide (right). The Self-Attention / CTC + LM result is with the same training data is also shown for comparison.}
\label{fig:cer_per_line_bucket}
\end{figure*}

\begin{figure}[t]
\begin{center}
  \includegraphics[width=0.95\linewidth]{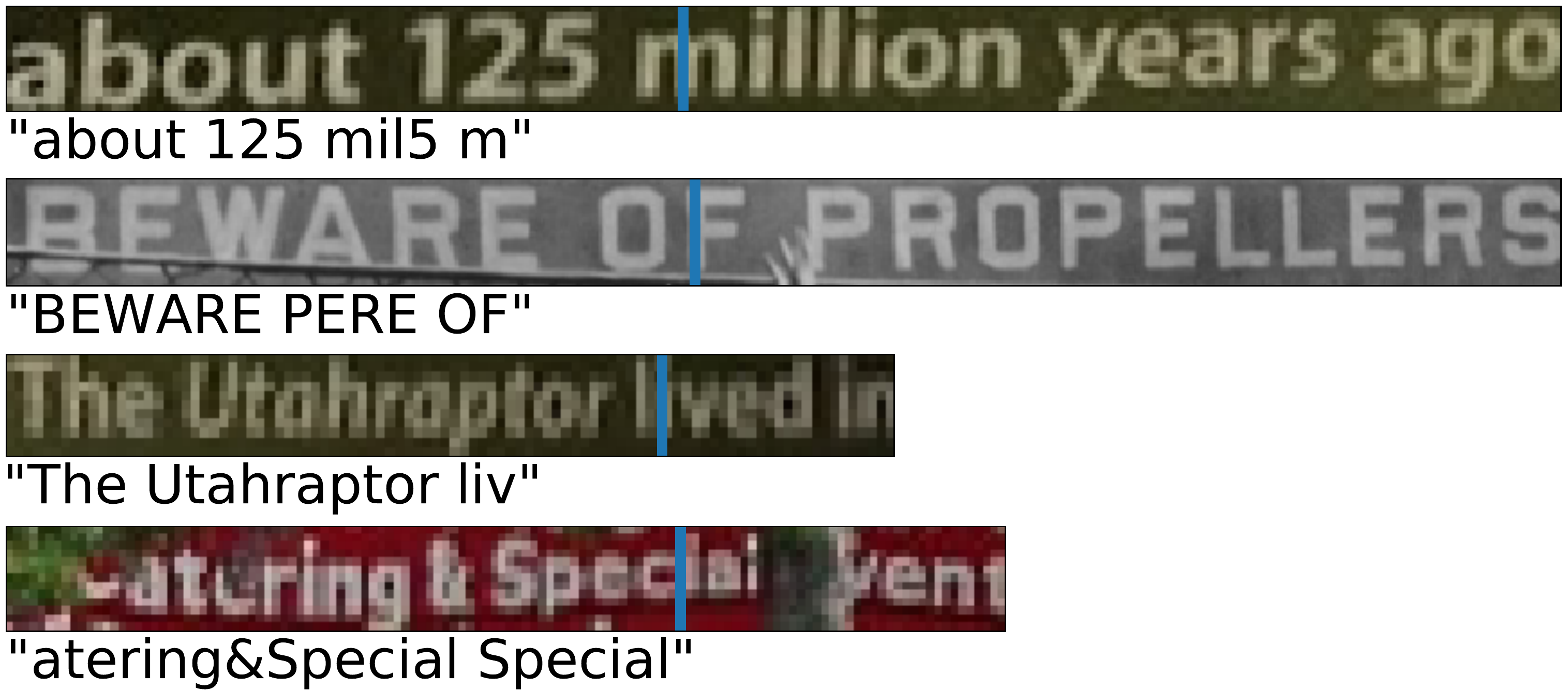}
\end{center}
  \caption{Failure cases for the Self-Attention / Transformer Decoder model, trained with images of maximum width 256 and evaluated with no resizing. Below each image, the TLR results. The vertical blue line denotes 256 pixel mark (height-normalized to 40 pixels)}
\label{fig:failure_modes_no_resize}
\end{figure}

The clear winner overall is the model that combines a Self-Attention encoder, the CTC decoder with LM, using 12 Self-Attention layers or more. Our absolute best run, with 16 Self-Attention layers, produces a 4\% relative CER improvement, with 8 million less parameters, over the second best model, a Self-Attention / Transformer Decoder with 20 encoder layers.

Different from other encoders, the BiLSTM encoder improved only slightly with more layers. This encoder is also more prone to overfitting and overall performs worse than the other two.

Without the explicit language model, CTC and the Transformer are overall competitive as decoders for the Self-Attention Module, but there are trade-offs. The CTC model has a better CER at the same number of parameters with similar WPA. However, the biggest Transformer model performs somewhat better in both metrics than the best CTC. Latency tells a different story. In Fig.~\ref{fig:latencies}, latencies for our non-recurrent models are represented for an input size of $40\times 320$. CTC is 5 to 10 times faster than the Transformer decoder while increasing the capacity of the encoder or adding a LM contributes only a small overhead

\subsection{Evaluation on Public Datasets}

In Table~\ref{tab:sota-comparison} we show the performance on the test sets of the public datasets of our best models trained on \emph{Synth + Public}, along with some of the best results reported in the literature. The table shows that in this training configuration, these models are competitive with SOTA, especially if one considers only models that do not include a rectification module. The Self-Attention/CTC model performs slightly worse than Self-Attention/Transformer in this configuration. Despite its small relative size, the inclusion of the \emph{Public} dataset is key to achieve this performance; training on synthetic data only yields significantly worse results.

We also show the evaluation results for the \emph{Internal} and \emph{Internal + Public} training runs, on the test sets of the public datasets for both decoders.  We stress that 
by presenting these results we do not intend to establish a comparison with SOTA, since the training data in this case differs from the public datasets typically used. No benefit was found from including the training portion of the scene-text public datasets in this training configuration. In fact, the runs that performed best on scene-text datasets corresponded to models trained with only \emph{Internal} data.
Conversely, adding the train sets of the handwritten public data have a substantial impact. This is most manifest for RIMES, in which the official training and test splits contain many examples with repeated labels.

\subsection{Resizing}
\label{resizing}

In order to train the Transformer decoder models on TPUs, images are first resized to a fixed height and then either padded or anisotropically shrunk to a fixed maximum width $M$. Reducing $M$ could in principle lead to faster training times, while potentially impacting performance, particularly for longer lines.

The performance changes with respect to $M$ are shown in Fig.~\ref{fig:cer_per_line_bucket}. In the figure we plot CER as a function of validation image width, grouped into buckets of 100 pixels. We considered two inference configurations differing on whether the validation images were resized to width $M$ or not. The Self-Attention/CTC + LM result, when trained on the same data, is shown for comparison. 

It is observed in Fig.~\ref{fig:cer_per_line_bucket} that the CERs drastically increase after $M$ without resizing in both max widths. Fig.~\ref{fig:failure_modes_no_resize} shows typical failure cases without resizing. The recognizer appears to essentially stop decoding at the maximum width, often outputting the same characters observed around $M$ repeatedly. Resizing the images to the maximum width at inference time alleviates the issue. However, there is a limitation with the approach. In our experiments, the accuracy started to noticeably degrade once the width of the image becomes longer than roughly $2 \times M$. In contrast, the accuracy of the CTC decoder is insensitive to the image width, which is another advantage of the CTC decoder.

\section{Conclusions}
\label{conclusions}

In this work we investigated the performance of representative encoder/decoder architectures as universal text-line recognizers. In the decoder comparison, we found that CTC, compounded with a language model, yielded the overall superior performance. In the absence of a LM, CTC and the Transformer are competitive, with CTC dominating in some cases (GRCL) and the Transformer in others (BiLSTM). On the other hand, in the encoder comparison, Self-Attention was the overall winner and both decoders were similarly accurate without LM. Interestingly, the unstudied Self-Attention/CTC + LM model, is our best. \cite{KaiChen:ICDAR2019} showed that attention-based decoders can still benefit from an external language model. Investigations on the effectiveness of external language models with transformer decoders will be future work. 

We also considered issues derived from the presence of long images in the distribution of examples. There are at least two new aspects that need to be considered, efficiency and performance. Long images affect the efficiency of models with the Self-Attention encoder due to quadratic scaling with image length. We showed that this problem can be solved for CTC models without performance loss by chunking the images. Training on images of fixed maximum width affects the performance on longer images of models that make use of the Transformer decoder. This problem can be alleviated, although not entirely eliminated, by resizing the images to the train width.

\newpage

{\small
\bibliographystyle{ieee_fullname}
\bibliography{main}
}

\end{document}